\documentclass[10pt,twocolumn,letterpaper]{article}

\usepackage{ijcb}
\usepackage{times}
\usepackage{epsfig}
\usepackage{graphicx}
\usepackage{amsmath}
\usepackage{amssymb}
\usepackage{subcaption}

\ijcbfinalcopy 

\newcommand{\paragraphb}[1]{\vspace{0.75ex}\noindent{\bf #1}\hspace*{.3em}}

\ifijcbfinal\fi

\makeatletter
\def\ps@IEEEtitlepagestyle{
\def\@oddfoot{\mycopyrightnotice}
\def\@evenfoot{}
}
\def\mycopyrightnotice{
}
\makeatother

\begin{document}

\title{Universal Adversarial Spoofing Attacks against Face Recognition}

\author{Takuma Amada\thanks{Equal contribution with authors in alphabetical order.}, Seng Pei Liew$^{*}$\thanks{Currently at LINE Corporation, Japan.}, Kazuya Kakizaki, Toshinori Araki\\
NEC Corporation\\
7-1, Shiba, 5-chome Minato-ku, Tokyo 108-8001 Japan\\
{\tt\small \{t-amada, kazuya1210, toshinori\_araki\}@nec.com,}
{\tt\small  sengpei.liew@gmail.com}
}

\maketitle
\thispagestyle{empty}

\begin{abstract}
We assess the vulnerabilities of deep face recognition systems for images that falsify/spoof multiple identities simultaneously.
We demonstrate that, by manipulating the deep feature representation extracted from a face image via imperceptibly small perturbations added at the pixel level using our proposed Universal Adversarial Spoofing Examples (UAXs), one can fool a face verification system into recognizing that the face image belongs to multiple different identities with a high success rate.
One characteristic of the UAXs crafted with our method is that they are universal (identity-agnostic); they are successful even against identities not known in advance.
For a certain deep neural network, we show that we are able to spoof almost all tested identities (99\%), including those not known beforehand (not included in training).
Our results indicate that a multiple-identity attack is a real threat and should be taken into account when deploying face recognition systems.

\end{abstract}
\let\thefootnote\relax\footnotetext{\mycopyrightnotice}

\section{Introduction}
\label{sec:intro}

Biometrics authentication is a security measure for uniquely identifying an individual on the basis of his/her physical or behavioral traits. 
Commonly used biometrics sources include face, fingerprint, and voice. 
Face recognition is common due to its high accuracy and seamless user interaction.

Advances in deep neural networks (DNNs) have greatly improved the performance of face recognition \cite{DBLP:conf/cvpr/TaigmanYRW14,DBLP:conf/nips/SunCWT14,DBLP:conf/bmvc/ParkhiVZ15}. For these and other reasons, face recognition is being used real-life applications, ranging from immigration inspection to smartphone authentication.

As with other forms of biometrics authentication, face authentication is susceptible to spoofing attacks, a type of attack in which an adversary attempts to bypass authentication by appearing to be genuine. 
There are mainly two approaches to bypassing authentication via spoofing. 
First, the adversary may carry out spoofing during authentication, e.g., present an image that looks genuine during authentication in real-time. Second, the adversary may register adversarial images in the system's database, i.e., \textit{data poisoning}, to confound the system during authentication occurring sometime later. We mainly focus on the latter approach in this study. 

Consider the following scenario in which an adversary registers/enrolls a slightly modified image of a genuine individual, the real identity of whom is denoted as \textit{A}, to the authentication system. 
During authentication, the system compares the enrolled image with a user-input image. 
We consider an attack to be successful only when the system is able to authenticate \textit{A} (no reasonable suspicion arises) as well as another individual \textit{B}, where $B\neq A$ (spoofing is successful). 
This allows \textit{B} to be authenticated even though \textit{B} is not enrolled, compromising the authentication system.

A more agnostic form of spoofing, where the modified image matches with more than two individuals, is also possible.
An adversary equipped with such an image can potentially break into multiple distinct authentication systems without needing knowledge about the specific identities registered to the authentication systems.

This led us to investigate \textit{universal} spoofing attacks against face recognition.
Particularly, we consider an adversary that registers an adversarial example (AX), an input image with a small amount of noise added, to an authentication system. 
Such an image with inconceivably small noise is preferred from the adversary's perspective, as it helps hide the adversary's trace without raising human suspicion.
We call such images that spoof multiple identities \textbf{universal adversarial spoofing examples (UAXs)}, to differentiate them from conventional AXs with which the only aim is to to cause misidentification.
The UAX matches with faces images of different genders, races, illumination, poses, and expressions.

The highlights of our contribution in this work are as follows:
\begin{itemize}
\item
Using adversarial perturbations, we propose a UAX-crafting method for enabling universal multiple-identity spoofing attacks on face recognition systems.

\item
Experimental results on multiple datasets and models indicate that our method is effective not only on targeted/known identities, but also generalizable to unknown identities.
In particular, the UAXs crafted with our method have a 99\% spoofing success rate (for both known and unknown identities we have tested them with) against the VGG2 feature extractor.

\end{itemize}

These vulnerabilities severely undermine the authentication capability of current face recognition systems and should be treated as a legitimate threat when deploying such systems in an industrial setting.

The rest of the paper is organized as follows. 
We first provide preliminaries and background relevant to our work in Section \ref{sec:pre}.
In section \ref{sec:method}, we present the proposed method of crafting UAXs. 
We discuss the extensive experimental evaluations we considered in Section \ref{sec:eval}.
Finally, we conclude the paper in Section \ref{sec:conclusion}.

\section{Background}
\label{sec:pre}

\paragraphb{Face recognition.} Broadly speaking, the task of face recognition is divided into two categories. The first category is face identification, or \textit{closed-set face recognition}, where the task is to execute a multi-class classification of faces belonging to a set of pre-determined identities. \footnote{Face recognition systems should reject unknown identities not belonging to the pre-determined set of identities. The \textit{open-set face recognition} protocol, which was designed to achieve this, has also been investigated \cite{DBLP:journals/corr/GuntherCRB17}.}

The second category is face verification. During authentication, the face recognition system compares the user-input image with images enrolled in the system's database and determines whether they belong to the same identity. 
The focus of our study was on face verification.

State-of-the-art face recognition systems use DNNs to extract a low-dimensional feature representation from a face image for face verification \cite{DBLP:conf/cvpr/LiuWYLRS17,DBLP:conf/cvpr/WangWZJGZL018,DBLP:conf/cvpr/DengGXZ19}. 
The similarity of two images can then be quantified by measuring the distance between their feature representations. 
Euclidean distance and cosine distance are common similarity metrics (the smaller the distance is, the more similar the images are).
The system subsequently determines if the two images belong to the same individual on the basis of a pre-determined decision threshold. 

The DNN architectures used to train face recognition feature extractors are typically those that work well in the image recognition domain. 
The VGG \cite{DBLP:journals/corr/SimonyanZ14a} and Inception \cite{DBLP:conf/cvpr/HeZRS16} networks are common DNN architectures.

We now describe the training process of feature extractors.
One common approach is training a DNN to execute classification on a pre-determined set of identities.
The output of the penultimate layer is then treated as the feature to be used for face verification.

Another approach involves training the feature extractor directly via metric learning \cite{DBLP:conf/cvpr/SchroffKP15}.
This approach involves constructing a triplet of two matching face images and a non-matching face image, but suffers from scalability issues, i.e., the number of triplet combinations increases exponentially with the number of training data points. 
We used feature extractors implementing the former approach in this study. 

It is worthwhile to note that recent releases of large annotated databases have also helped advance the field of face recognition.
Widely used databases include CASIA-WebFace \cite{DBLP:journals/corr/YiLLL14a}, VGGFace2 \cite{DBLP:conf/fgr/CaoSXPZ18}, MS1M \cite{DBLP:conf/eccv/GuoZHHG16}, and LFW \cite{LFWTech}.
The CASIA-WebFace dataset contains 0.49M images from 10,575 celebrities. 
The VGGFace2 is a large-scale dataset with large variations in pose, age, illumination, ethnicity, and profession. A total of 3.31M images from 9131 celebrities can be found in this dataset. 
The MS-Celeb1M dataset (MS1M) is another large-scale dataset with about 10k celebrities with 10M images. 
A refined version of this dataset (MS1MV2) containing less noise and with 3.8M images of 85,164 celebrities is also available \cite{DBLP:conf/cvpr/DengGXZ19}.
The Labeled Face in the Wild dataset (LFW) is a smaller dataset (13k images from 1,680 people) that is typically used for validation.

\paragraphb{Adversarial examples.} The phenomenon in which a small crafted perturbation (noise) added to an input may lead to misclassification by the DNN was first discussed in \cite{DBLP:journals/corr/SzegedyZSBEGF13}. 
Subsequently, various methods of crafting adversarial perturbations have been proposed, 
including those involving gradient updates \cite{DBLP:journals/corr/GoodfellowSS14,DBLP:conf/iclr/KurakinGB17,DBLP:conf/iclr/MadryMSTV18}. For example, a ``fast" one-step gradient update along the direction of the sign of gradient (FGSM) was carried out \cite{DBLP:journals/corr/GoodfellowSS14}. Iterative methods with higher probability of fooling DNNs have been proposed in \cite{DBLP:conf/iclr/KurakinGB17}. 
The Carlini-Wagner attack relies on optimization problem solving \cite{DBLP:conf/sp/Carlini017}, which has been empirically shown to be extremely successful at attacking DNNs with minimal perturbation.

While the aforementioned methods focus on causing misclassification, Rozsa et al.  \cite{DBLP:conf/icb/RozsaGB17} crafted AXs that align the internal layer representation with the target representation, which can be adapted to generate AXs to manipulate the feature representation of face images when one considers attacks on face verification.

We also notice that the study of ``universal'' adversarial perturbation is available in the literature \cite{DBLP:conf/cvpr/Moosavi-Dezfooli17}, where the aim of such an attack is to create a single perturbation to mis-classify various images (e.g., images from class A mis-classified as class B).
The attacks crafted with our method are different in the sense that, when the perturbation is added to a face image, they are able to match many classes (e.g., class A mis-classified not only as class B, but also class C, class D etc.).
Since previous works focused on classification problems, it is not obvious that such universal behavior exists in face verification (where the task is calculating how similar two feature vectors are), which is rather different from classification (where the task is calculating how likely an image belongs to a certain class).

\paragraphb{Attacks on face recognition.} Security issues of face recognition have attracted much attention.
For example, Sharif et al. \cite{DBLP:conf/ccs/SharifBBR16} showed that it is possible to fool face recognition systems deployed in the physical world by adding perturbations in the eyeglass region.
In \cite{DBLP:conf/cvpr/DongSWLL0019}, a method of attacking face recognition in a black-box fashion has been proposed.
Attacking face recognition using generative adversarial networks (GANs) has been studied as well \cite{DBLP:journals/corr/abs-1811-12026}.

It should be noted that most studies on attacks on face recognition focus on tricking the authentication system into misidentifying an individual, orthogonal to our adversarial purposes of spoofing multiple identities.  

Finding a generic sample that is similar to many of the enrolled templates is known as a wolf attack in biometrics \cite{DBLP:conf/icb/UneOI07}. \footnote{It is also known as dictionary attack in computer security.}
Previous works have focused on finger-vein-, fingerprint- and voice-based authentication systems \cite{DBLP:conf/icb/UneOI07,DBLP:conf/btas/BontragerRTMR18,DBLP:conf/interspeech/MarrasKMF19}.

 \cite{DBLP:journals/corr/abs-1906-08507} and \cite{DBLP:conf/icb/NguyenYEM20} investigated multiple-identity attacks and have adversarial purposes similar to the present work. To carry out multiple-identity attack, \cite{DBLP:journals/corr/abs-1906-08507} used face morphing methods and searched in a gallery for natural faces that match with two identities. However, they considered neither adversarial perturbations nor dictionary attacks, where a single face can match with more than two identities, which constitute the core part of the present work, were considered in \cite{DBLP:journals/corr/abs-1906-08507}.
\cite{DBLP:conf/icb/NguyenYEM20} considers \textit{Master Faces}, multi-identity spoofing images crafted using GANs. This is different from our method of crafting images using adversarial perturbations, which is a stealthier form of attack.
Our study is more extensive in the sense that we demonstrated the effectiveness of our attack by performing evaluation on multiple DNNs, in contrast with \cite{DBLP:conf/icb/NguyenYEM20}, who conducted an evaluation only on a single DNN.

\section{Proposed Method}
\label{sec:method}

Our aim is to craft a UAX that can spoof as many identities as possible.
We craft the UAX by adding perturbations to a ``seed" genuine image, $x_A$.
Given a training dataset, $A_{\rm train}$, our crafting strategy is as follows. We add perturbations to $x_A$ such that its similarities with \textit{all} images from $A_{\rm train}$ are maximized.
We further prepare a separate dataset containing a disjoint population of individuals, $A_{\rm test}$, to test the spoofing capability of UAXs with unseen identities.

\subsection{Formulating Universal Adversarial Spoofing Examples}
\label{subsec:targetform}
Let us provide a more precise formulation of our proposed method.
Let $\phi(\cdot)$ denote the DNN feature extractor. 
The feature representation extracted from the image of individual $A$, $x_A$, and individual $B$, $x_B$, are $\phi(x_A)$ and $\phi_(x_B)$, respectively. 
Without loss of generality, we consider adding a small perturbation, $\nu$, to $x_A$ so that the UAX may be written as 
\begin{align}
    x' \equiv x_A + \nu.
\end{align}
The corresponding feature representation $\phi(x')$ is required to have a large enough similarity score with $B$ for successful spoofing. 

Requiring $\nu$ to be small can be formulated as $||\nu||_p < \xi$, where $\xi$ is a parameter controlling the size of the perturbation, and $p$ refers to the $l_p$ norm with $p \in [1,\infty)$ (we focus on $p=\infty$ for the rest of this paper).

 \subsection{Algorithm}
 The goal with proposed method is to craft a UAX ($x'$) that spoofs as many images in $A_{\rm train}$ as possible. 
 To achieve this, our algorithm iteratively searches for $x'$ that minimizes 
 \begin{align}
f(x';x^j) \equiv \left\| \phi(x') - \phi(x^j)\right\|_2
\end{align}
by searching for $\Delta \nu$ and aggregating it to the current value of $\nu$ at each iteration, where $x^j \in A_{\rm train}$.
 
The proposed method requires updating the gradient in a mini-batch fashion.
 We first initialize 
  \begin{align*}
   \nu &\leftarrow 0, \\
 x' &\leftarrow x_A + \nu.
 \end{align*}
At each iteration $i$, we then perform the following steps:
\begin{itemize}
\item
\textbf{Mini-batch sampling.} Sample a mini-batch of size $n$ from $A_{\rm train}$, i.e.,  $S^i_{\rm batch} \leftarrow \{x_B: x_B \in A_{\rm train}\}$.

\item 
\textbf{Loss evaluation and optimization.} Minimize 
\begin{align}
\label{eq:optim}
F(x';S^i_{\rm batch}) \equiv \frac{1}{n} \sum_{x_B\in S^i_{\rm batch}} \left\|\phi(x') - \phi(x_B)\right\|_2
\end{align}
under the constraint
\begin{align*}
\left\|\nu\right\|_p < \xi.
\end{align*}

\end{itemize}
The iteration may be terminated after a certain pre-determined number of iterations.
In this work, we test with 500 and 5,000 iterations.

\paragraphb{Optimization details.}
Before solving Eq. \ref{eq:optim}, we normalize the range of $x$ to $[0,1]$ . The following change from $x$ to $w$ is then made:
\begin{align}
x = \frac{1}{2} ( \text{tanh}\ (w) +1).
\end{align}
We solve for $w$ using the stochastic gradient descent (SGD) method.
Finally, we clip the perturbation ($\nu \leftarrow \mathcal{P}(\nu; p, \xi)$) using the following projection operator:
\begin{align}
\label{eq:project}
    &\mathcal{P}(\nu; p, \xi) \equiv  \operatorname*{argmin}_\rho \left\|\rho - \nu\right\|_2 \ \text{s.t.}\ ||\rho||_p < \xi. 
\end{align}
\section{Evaluation}
\label{sec:eval}
In this Section, we describe the experimental evaluation of the proposed  method. 
Two attack scenarios are considered: 
\begin{itemize}
    \item \textit{White-box attacks.} The attacker crafts UAXs using the victim's feature extractor. 
    \item \textit{Black-box attacks.} The attacker crafts UAXs using feature extractors different from the one he or she is targeting.
\end{itemize}

\subsection{Experimental Setup}
\paragraphb{Network settings.}
To test the validity of our approach, we conducted an empirical evaluation by using a variety of open-source feature extractors as the target of our attack, namely VGG \footnote{https://github.com/yzhang559/vgg-face}, VGG2 \footnote{https://github.com/rcmalli/keras-vggface}, SphereFace \footnote{https://github.com/clcarwin/sphereface\_pytorch}, and InsightFace \footnote{https://github.com/TreB1eN/InsightFace\_Pytorch}.

These feature extractors use well-known backbone architectures (convolutional neural networks (CNNs) and Resnets) and have been trained to classify a pre-determined set of identities in a supervised manner.
Details of the network settings of the feature extractors, including training method, architecture and dataset used for training are summarized in Table \ref{tab:fe}.

\begin{table}[t] 
\scriptsize
\centering
\caption{Details of feature extractors used in this work. Name of the feature extractor, training method (loss function used), DNN architecture and dataset used for training are shown. See text for links to open-source codes.}
\label{tab:fe}
\begin{tabular}{|c|c|c|c|c|}
\hline 
 Model name & Training method & Architecture & Training dataset  \\ 
\hline 
\hline 
VGG   & Softmax &  VGG16&  VGGFace\\
\hline 
VGG2    &  Softmax &  Resnet50&  VGGFace2\\
\hline 
SphereFace   & SphereFace \cite{DBLP:conf/cvpr/LiuWYLRS17}& 36-layer CNN& CASIA-WebFace   \\
\hline 
InsightFace   & ArcFace \cite{DBLP:conf/cvpr/DengGXZ19}& SE-Inception50 & MS1MV2 \cite{DBLP:conf/cvpr/DengGXZ19}  \\
\hline
\end{tabular}
\end{table}

\begin{table}[t] 

\centering
\caption{Datasets and number of instances used for training and testing UAXs.}
\label{tab:dataset}
\begin{tabular}{|c|c|c|}
\hline 
 Dataset name &   Train & Test  \\ 
\hline 
\hline 
VGGFace2   & 10,000 &  5,000  \\
\hline 
LFW    &  3,438 &  1,549\\
\hline 
\end{tabular}
\end{table}

\paragraphb{Datasets.}
Four datasets were constructed from VGGFace2 and LFW datasets to craft (train) and evaluate (test) UAXs (two each for training and testing). From VGGFace2 (LFW) dataset, we selected 10 (5) images to be the ``seeds" that will be added with perturbations to craft UAXs.

The VGGFace2 train (test) dataset used for evaluation was constructed by sampling 2000 (1000) identities and 5 images from each identity from the train (test) subset of VGGFace2.

To evaluate with LFW dataset, we selected all images from the recommended train and test pairs \cite{LFWTech} to construct the train (3,438 images) and test (1,549 images) dataset respectively.

Note that the train and test datasets were mutually exclusive (no over-lapping identities). 
The details of the datasets are summarized in Table \ref{tab:dataset}.

For image pre-processing, we employed the multitask convolutional neural network (MTCNN) to detect and crop face images to $112 \times 112$ pixels \cite{DBLP:journals/spl/ZhangZLQ16}. 

\paragraphb{Evaluation metrics.}
Before deployment, practitioners of the face recognition systems need to determine a decision threshold on the similarity score to verify whether two face images belong to the same identity.
In this study, we used the equal error rate (EER) as the decision threshold.
This measurement is also known as the imposter attack presentation match rate (IAPMR). 

The discriminative power of a feature extractor is measured using the false matching rate (FMR) (the smaller an FMR is, the better a feature extractor is).
Conversely, the FMR is also used as a metric to quantify the attack success rate (the higher an FMR is, the higher attack success rate a UAX has).

\subsection{White-box attacks}
\label{subsec:exp2}
For our first evaluation, we assumed that the adversary has full knowledge of the victim's feature extractor, such that a white-box attack is viable, where the adversary crafts the UAXs using the victim's feature extractor.
We fixed the perturbation size to $\xi = 10 / 255$, or $\epsilon = 10$.

By crafting the UAXs on the basis of the discussion in Section \ref{sec:method}, we show how the FMRs change from non-UAXs (zero-effort imposters) to UAXs.
Tables \ref{tab:res1}-\ref{tab:res4} list the results for the four targeted feature extractors.
Note that the FMRs are shown as the average among the 10 (5) UAXs we crafted on the basis of the VGGFace2 (LFW) dataset.

The UAXs consistently outperformed the non-UAXs by a significant margin at spoofing.
The VGG2 feature extractor was the most vulnerable; 99\% of the images in the train and test datasets were successfully spoofed since the UAXs were crafted with 5,000 iterations.
Even against the state-of-the-art feature extractor, InsightFace, the UAXs were relatively successful, being able to spoof as much as around 20 \% of the dataset (for both train and test datasets).

\begin{figure}[!tbp]
\scriptsize
  \begin{subfigure}[b]{0.2\textwidth}
    \includegraphics[width=4.0cm]{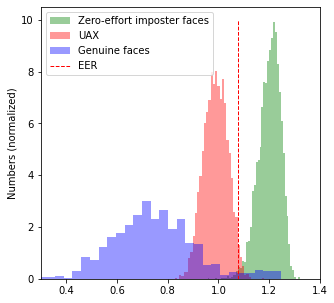}
    \caption{Train}
    \label{fig:m1}
  \end{subfigure}
  \hspace{5mm}
  \begin{subfigure}[b]{0.2\textwidth}
    \includegraphics[width=4.0cm]{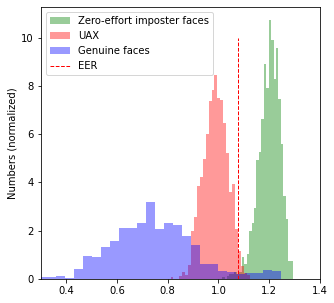}
    \caption{Test}
    \label{fig:m2}
  \end{subfigure}
  \caption{Euclidean distance distributions of LFW face images with respect to train and test datasets. Distance distributions of same-identity face images (\textit{genuine faces}), different-identity face images (\textit{zero-effort imposter faces}), and UAXs without adversarial perturbations (\textit{UAXs}).}
      \label{fig:masterface}
\end{figure}

Figure \ref{fig:masterface} gives another view of UAXs, where the score distributions of genuine faces, zero-effort imposter faces, and UAXs are shown.
Zero-effort imposter faces consisted of ``seed" images and different identities as we use in white-box evaluation.
Genuine faces were constructed by sampling pairs of face images from the same identity.

Although the adversarial perturbation was quite small, each sample could spoof over 99\% of the different identities.

\noindent \textbf{Observations.}
The feature vector of our face recognition module is of 512 dimensions, and the hypersphere is theoretically large enough to hold large-scale identities with a small misidentifying probability if the feature vectors of each identity are distributed uniformly (see, e.g., \cite{DBLP:conf/cvpr/DengGXZ19}). 
Our results indicate otherwise. 
It seems that the feature vectors of different identities are more “concentrated” than expected.
We hope this observation spurs further theoretical investigation and analysis in the future.
\begin{table}[t] 
\scriptsize
\centering
\caption{Average FMRs for ``seed" images (*-*-$b$) and UAXs (*-*-$a$, in bold text), when measured with instances of train (*-$train$-*) and test (*-$test$-*) datasets, constructed from VGGFace2 ($VGG$-*-*) and LFW ($LFW$-*-*). The \textit{VGG} feature extractor was used in this configuration.}
\label{tab:res1}
\begin{tabular}{|l|c|c|}
\hline 
 Dataset & $\epsilon=10$, $500$ iteration &$\epsilon=10$, $5,000$ iteration  \\ 
\hline 
 VGG-train-b   & $0.094\pm 0.033$& $0.094\pm 0.033$  \\
\textbf{VGG-train-a}   & $\bf 0.15\pm 0.071$& $\bf 0.32\pm 0.11$  \\
\hline 
VGG-test-b   & $0.098\pm 0.033$& $0.098 \pm 0.033$  \\
\textbf{VGG-test-a}   & $\bf 0.15\pm 0.061$& $\bf 0.32\pm 0.11$  \\
\hline 
LFW-train-b   & $0.061\pm 0.018$& $0.061\pm 0.018$  \\
\textbf{LFW-train-a}   & $\bf 0.15\pm 0.032$& $\bf 0.34\pm0.078$  \\
\hline 
LFW-test-b   & $0.063\pm 0.018 $& $0.063\pm 0.018 $  \\
\textbf{LFW-test-a}   & $\bf 0.15\pm 0.031$& $\bf 0.34\pm 0.083$  \\
\hline
\end{tabular}
\end{table}

\begin{table}[t] 
\scriptsize
\centering
\caption{VGG2 feature extractor was used in this configuration. See title of Table \ref{tab:res1} for details.} 
\label{tab:res2}
\begin{tabular}{|l|c|c|}
\hline 
 Dataset & $\epsilon=10$, $500$ iteration &$\epsilon=10$, $5,000$ iteration  \\ 
\hline 
 VGG-train-b   & $0.056\pm 0.030$& $0.056\pm 0.030$  \\
\textbf{VGG-train-a}   & $\bf 0.69\pm 0.18$& $\bf 0.99\pm 0.0017$  \\
\hline 
VGG-test-b   & $0.060\pm 0.029$& $0.060\pm 0.029$  \\
\textbf{VGG-test-a}   & $\bf 0.70\pm 0.17$& $\bf 0.99\pm 0.0018$  \\
\hline 
LFW-train-b   & $0.068\pm 0.023$& $0.068\pm 0.023$  \\
\textbf{LFW-train-a}   & $\bf 0.76\pm 0.047$& $\bf 0.99\pm0.0063$  \\
\hline 
LFW-test-b   & $0.066\pm 0.027 $& $0.066\pm 0.027 $  \\
\textbf{LFW-test-a}   & $\bf 0.77\pm 0.049$& $\bf 0.99\pm0.0054$  \\
\hline
\end{tabular}
\end{table}

\begin{table}[t] 
\scriptsize
\centering
\caption{Insightface feature extractor was used in this configuration. See title of Table \ref{tab:res1} for details.} 
\label{tab:res3}
\begin{tabular}{|l|c|c|}
\hline 
 Dataset & $\epsilon=10$, $500$ iteration &$\epsilon=10$, $5,000$ iteration  \\ 
\hline 
 VGG-train-b   & $0.075\pm 0.013$& $0.075\pm 0.013$  \\
\textbf{VGG-train-a}   & $\bf 0.19\pm 0.025$& $\bf 0.25\pm 0.012$  \\
\hline 
VGG-test-b   & $0.075\pm 0.011$& $ 0.075\pm 0.011 $  \\
\textbf{VGG-test-a}   & $\bf 0.18\pm 0.019$& $\bf 0.23\pm 0.014$  \\
\hline 
LFW-train-b   & $0.056\pm 0.011$& $0.056\pm 0.011$  \\
\textbf{LFW-train-a}   & $\bf 0.20\pm 0.013$& $\bf 0.24\pm0.00051$  \\
\hline 
LFW-test-b   & $0.057\pm 0.0040 $& $0.057\pm 0.0040 $  \\
\textbf{LFW-test-a}   & $\bf 0.20\pm 0.014$& $\bf 0.25\pm 0.0017$  \\
\hline
\end{tabular}
\end{table}

\begin{table}[t] 
\scriptsize
\centering
\caption{SphereFace feature extractor was used in this configuration. See title of Table \ref{tab:res1} for details.} 
\label{tab:res4}
\begin{tabular}{|l|c|c|}
\hline 
 Dataset & $\epsilon=10$, $500$ iteration &$\epsilon=10$, $5,000$ iteration  \\ 
\hline 
 VGG-train-b   & $0.11\pm 0.034$& $0.11\pm 0.034$  \\
\textbf{VGG-train-a}   & $\bf 0.60\pm 0.040$& $\bf 0.63\pm 0.0032$  \\
\hline 
VGG-test-b   & $0.11\pm 0.028 $& $ 0.11\pm 0.028$  \\
\textbf{VGG-test-a}   & $\bf 0.62\pm 0.040$& $\bf 0.66\pm 0.0022$  \\
\hline 
LFW-train-b   & $0.067\pm0.038$& $0.067\pm0.038$  \\
\textbf{LFW-train-a}   & $\bf 0.56\pm 0.0011$& $\bf 0.57\pm0.00024$  \\
\hline 
LFW-test-b   & $0.067\pm 0.038 $& $0.067\pm 0.038 $  \\
\textbf{LFW-test-a}   & $\bf 0.57\pm 0.0024$& $\bf 0.58\pm 0.000072$  \\
\hline
\end{tabular}
\end{table}

\subsection{Black-box attacks}
\label{subsec:black}
\begin{figure}[!tbp]
\scriptsize
  \begin{subfigure}[b]{0.2\textwidth}
    \includegraphics[width=4.0cm]{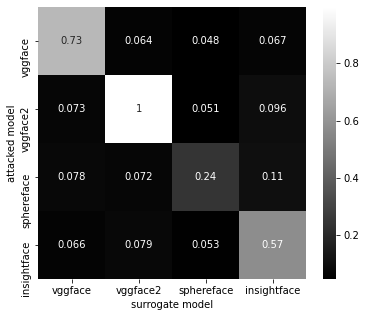}
    \caption{Train}
    \label{fig:m1}
  \end{subfigure}
  \hspace{5mm}
  \begin{subfigure}[b]{0.2\textwidth}
    \includegraphics[width=4.0cm]{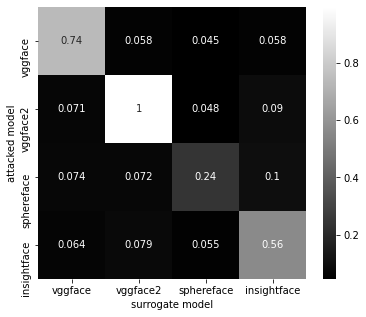}
    \caption{Test}
    \label{fig:m2}
  \end{subfigure}
  \caption{Adversarial transferability among VGG, VGG2, InsightFace, SphereFace models. We tested FMRs on UAXs with 5,000 iterations, $\epsilon$ is 10.}
      \label{fig:masterface_bb}
\end{figure}

For the second evaluation, we transferred the UAXs generated from each feature extractor to the other ones and measured the FMRs for all combinations of different feature extractors.
Figure \ref{fig:masterface_bb} shows the heat maps of adversarial transferability in the train and test datasets.
The UAXs generally did not transfer well to the other feature extractors.
In the best case, $10\%$ of the UAXs crafted by attacking the InsightFace model were transferable to the SphereFace model.  

\section{Discussion and conclusion}
\label{sec:conclusion}

We conducted the first analysis of multiple-identity attacks on face verification using adversarial perturbations.
We showed, particularly in a white-box attack scenario, that an adversary can significantly enhance the spoofing capability of face images using our proposed method. Even without the knowledge of the identities registered to a certain face recognition system, an adversary can use the universal spoofing capability of the image to break into the authentication process with a markedly improved success rate.

Our crafted UAXs do not transfer well, as shown in Section \ref{subsec:black}. 
Some direct methods resolving this include training our UAXs with many different architectures to increase transferability, which is left for future work.
Other future research directions include developing defenses against UAXs.

{\small
\bibliographystyle{ieee}
\bibliography{refs}

\begin{thebibliography}{10}\itemsep=-1pt

\bibitem{DBLP:journals/corr/abs-1906-08507}
J.~T.~A. Andrews, T.~Tanay, and L.~D. Griffin.
\newblock Multiple-identity image attacks against face-based identity
  verification.
\newblock {\em CoRR}, abs/1906.08507, 2019.

\bibitem{DBLP:conf/btas/BontragerRTMR18}
P.~Bontrager, A.~Roy, J.~Togelius, N.~D. Memon, and A.~Ross.
\newblock Deepmasterprints: Generating masterprints for dictionary attacks via
  latent variable evolution\({}^{\mbox{*}}\).
\newblock In {\em 9th {IEEE} International Conference on Biometrics Theory,
  Applications and Systems, {BTAS} 2018, Redondo Beach, CA, USA, October 22-25,
  2018}, pages 1--9. {IEEE}, 2018.

\bibitem{DBLP:conf/fgr/CaoSXPZ18}
Q.~Cao, L.~Shen, W.~Xie, O.~M. Parkhi, and A.~Zisserman.
\newblock Vggface2: {A} dataset for recognising faces across pose and age.
\newblock In {\em 13th {IEEE} International Conference on Automatic Face {\&}
  Gesture Recognition, {FG} 2018, Xi'an, China, May 15-19, 2018}, pages 67--74.
  {IEEE} Computer Society, 2018.

\bibitem{DBLP:conf/sp/Carlini017}
N.~Carlini and D.~A. Wagner.
\newblock Towards evaluating the robustness of neural networks.
\newblock In {\em {IEEE} Symposium on Security and Privacy}, pages 39--57.
  {IEEE} Computer Society, 2017.

\bibitem{DBLP:conf/cvpr/DengGXZ19}
J.~Deng, J.~Guo, N.~Xue, and S.~Zafeiriou.
\newblock Arcface: Additive angular margin loss for deep face recognition.
\newblock In {\em {CVPR}}, pages 4690--4699. Computer Vision Foundation /
  {IEEE}, 2019.

\bibitem{DBLP:conf/cvpr/DongSWLL0019}
Y.~Dong, H.~Su, B.~Wu, Z.~Li, W.~Liu, T.~Zhang, and J.~Zhu.
\newblock Efficient decision-based black-box adversarial attacks on face
  recognition.
\newblock In {\em {IEEE} Conference on Computer Vision and Pattern Recognition,
  {CVPR} 2019, Long Beach, CA, USA, June 16-20, 2019}, pages 7714--7722.
  Computer Vision Foundation / {IEEE}, 2019.

\bibitem{DBLP:journals/corr/GoodfellowSS14}
I.~J. Goodfellow, J.~Shlens, and C.~Szegedy.
\newblock Explaining and harnessing adversarial examples.
\newblock In Y.~Bengio and Y.~LeCun, editors, {\em 3rd International Conference
  on Learning Representations, {ICLR} 2015, San Diego, CA, USA, May 7-9, 2015,
  Conference Track Proceedings}, 2015.

\bibitem{DBLP:journals/corr/GuntherCRB17}
M.~G{\"{u}}nther, S.~Cruz, E.~M. Rudd, and T.~E. Boult.
\newblock Toward open-set face recognition.
\newblock {\em CoRR}, abs/1705.01567, 2017.

\bibitem{DBLP:conf/eccv/GuoZHHG16}
Y.~Guo, L.~Zhang, Y.~Hu, X.~He, and J.~Gao.
\newblock Ms-celeb-1m: {A} dataset and benchmark for large-scale face
  recognition.
\newblock In B.~Leibe, J.~Matas, N.~Sebe, and M.~Welling, editors, {\em
  Computer Vision - {ECCV} 2016 - 14th European Conference, Amsterdam, The
  Netherlands, October 11-14, 2016, Proceedings, Part {III}}, volume 9907 of
  {\em Lecture Notes in Computer Science}, pages 87--102. Springer, 2016.

\bibitem{DBLP:conf/cvpr/HeZRS16}
K.~He, X.~Zhang, S.~Ren, and J.~Sun.
\newblock Deep residual learning for image recognition.
\newblock In {\em 2016 {IEEE} Conference on Computer Vision and Pattern
  Recognition, {CVPR} 2016, Las Vegas, NV, USA, June 27-30, 2016}, pages
  770--778. {IEEE} Computer Society, 2016.

\bibitem{LFWTech}
G.~B. Huang, M.~Ramesh, T.~Berg, and E.~Learned-Miller.
\newblock Labeled faces in the wild: A database for studying face recognition
  in unconstrained environments.
\newblock Technical Report 07-49, University of Massachusetts, Amherst, October
  2007.

\bibitem{DBLP:conf/iclr/KurakinGB17}
A.~Kurakin, I.~J. Goodfellow, and S.~Bengio.
\newblock Adversarial machine learning at scale.
\newblock In {\em 5th International Conference on Learning Representations,
  {ICLR} 2017, Toulon, France, April 24-26, 2017, Conference Track
  Proceedings}. OpenReview.net, 2017.

\bibitem{DBLP:conf/cvpr/LiuWYLRS17}
W.~Liu, Y.~Wen, Z.~Yu, M.~Li, B.~Raj, and L.~Song.
\newblock Sphereface: Deep hypersphere embedding for face recognition.
\newblock In {\em {CVPR}}, pages 6738--6746. {IEEE} Computer Society, 2017.

\bibitem{DBLP:conf/iclr/MadryMSTV18}
A.~Madry, A.~Makelov, L.~Schmidt, D.~Tsipras, and A.~Vladu.
\newblock Towards deep learning models resistant to adversarial attacks.
\newblock In {\em 6th International Conference on Learning Representations,
  {ICLR} 2018, Vancouver, BC, Canada, April 30 - May 3, 2018, Conference Track
  Proceedings}. OpenReview.net, 2018.

\bibitem{DBLP:conf/interspeech/MarrasKMF19}
M.~Marras, P.~Korus, N.~D. Memon, and G.~Fenu.
\newblock Adversarial optimization for dictionary attacks on speaker
  verification.
\newblock In G.~Kubin and Z.~Kacic, editors, {\em Interspeech 2019, 20th Annual
  Conference of the International Speech Communication Association, Graz,
  Austria, 15-19 September 2019}, pages 2913--2917. {ISCA}, 2019.

\bibitem{DBLP:conf/cvpr/Moosavi-Dezfooli17}
S.~Moosavi{-}Dezfooli, A.~Fawzi, O.~Fawzi, and P.~Frossard.
\newblock Universal adversarial perturbations.
\newblock In {\em 2017 {IEEE} Conference on Computer Vision and Pattern
  Recognition, {CVPR} 2017, Honolulu, HI, USA, July 21-26, 2017}, pages 86--94.
  {IEEE} Computer Society, 2017.

\bibitem{DBLP:conf/icb/NguyenYEM20}
H.~H. Nguyen, J.~Yamagishi, I.~Echizen, and S.~Marcel.
\newblock Generating master faces for use in performing wolf attacks on face
  recognition systems.
\newblock In {\em 2020 {IEEE} International Joint Conference on Biometrics,
  {IJCB} 2020, Houston, TX, USA, September 28 - October 1, 2020}, pages 1--10.
  {IEEE}, 2020.

\bibitem{DBLP:conf/bmvc/ParkhiVZ15}
O.~M. Parkhi, A.~Vedaldi, and A.~Zisserman.
\newblock Deep face recognition.
\newblock In X.~Xie, M.~W. Jones, and G.~K.~L. Tam, editors, {\em Proceedings
  of the British Machine Vision Conference 2015, {BMVC} 2015, Swansea, UK,
  September 7-10, 2015}, pages 41.1--41.12. {BMVA} Press, 2015.

\bibitem{DBLP:conf/icb/RozsaGB17}
A.~Rozsa, M.~G{\"{u}}nther, and T.~E. Boult.
\newblock {LOTS} about attacking deep features.
\newblock In {\em {IJCB}}, pages 168--176. {IEEE}, 2017.

\bibitem{DBLP:conf/cvpr/SchroffKP15}
F.~Schroff, D.~Kalenichenko, and J.~Philbin.
\newblock Facenet: {A} unified embedding for face recognition and clustering.
\newblock In {\em {IEEE} Conference on Computer Vision and Pattern Recognition,
  {CVPR} 2015, Boston, MA, USA, June 7-12, 2015}, pages 815--823. {IEEE}
  Computer Society, 2015.

\bibitem{DBLP:conf/ccs/SharifBBR16}
M.~Sharif, S.~Bhagavatula, L.~Bauer, and M.~K. Reiter.
\newblock Accessorize to a crime: Real and stealthy attacks on state-of-the-art
  face recognition.
\newblock In E.~R. Weippl, S.~Katzenbeisser, C.~Kruegel, A.~C. Myers, and
  S.~Halevi, editors, {\em Proceedings of the 2016 {ACM} {SIGSAC} Conference on
  Computer and Communications Security, Vienna, Austria, October 24-28, 2016},
  pages 1528--1540. {ACM}, 2016.

\bibitem{DBLP:journals/corr/SimonyanZ14a}
K.~Simonyan and A.~Zisserman.
\newblock Very deep convolutional networks for large-scale image recognition.
\newblock In Y.~Bengio and Y.~LeCun, editors, {\em 3rd International Conference
  on Learning Representations, {ICLR} 2015, San Diego, CA, USA, May 7-9, 2015,
  Conference Track Proceedings}, 2015.

\bibitem{DBLP:journals/corr/abs-1811-12026}
Q.~Song, Y.~Wu, and L.~Yang.
\newblock Attacks on state-of-the-art face recognition using attentional
  adversarial attack generative network.
\newblock {\em CoRR}, abs/1811.12026, 2018.

\bibitem{DBLP:conf/nips/SunCWT14}
Y.~Sun, Y.~Chen, X.~Wang, and X.~Tang.
\newblock Deep learning face representation by joint
  identification-verification.
\newblock In {\em {NIPS}}, pages 1988--1996, 2014.

\bibitem{DBLP:journals/corr/SzegedyZSBEGF13}
C.~Szegedy, W.~Zaremba, I.~Sutskever, J.~Bruna, D.~Erhan, I.~J. Goodfellow, and
  R.~Fergus.
\newblock Intriguing properties of neural networks.
\newblock In Y.~Bengio and Y.~LeCun, editors, {\em 2nd International Conference
  on Learning Representations, {ICLR} 2014, Banff, AB, Canada, April 14-16,
  2014, Conference Track Proceedings}, 2014.

\bibitem{DBLP:conf/cvpr/TaigmanYRW14}
Y.~Taigman, M.~Yang, M.~Ranzato, and L.~Wolf.
\newblock Deepface: Closing the gap to human-level performance in face
  verification.
\newblock In {\em {CVPR}}, pages 1701--1708. {IEEE} Computer Society, 2014.

\bibitem{DBLP:conf/icb/UneOI07}
M.~Une, A.~Otsuka, and H.~Imai.
\newblock Wolf attack probability: {A} new security measure in biometric
  authentication systems.
\newblock In S.~Lee and S.~Z. Li, editors, {\em Advances in Biometrics,
  International Conference, {ICB} 2007, Seoul, Korea, August 27-29, 2007,
  Proceedings}, volume 4642 of {\em Lecture Notes in Computer Science}, pages
  396--406. Springer, 2007.

\bibitem{DBLP:conf/cvpr/WangWZJGZL018}
H.~Wang, Y.~Wang, Z.~Zhou, X.~Ji, D.~Gong, J.~Zhou, Z.~Li, and W.~Liu.
\newblock Cosface: Large margin cosine loss for deep face recognition.
\newblock In {\em {CVPR}}, pages 5265--5274. {IEEE} Computer Society, 2018.

\bibitem{DBLP:journals/corr/YiLLL14a}
D.~Yi, Z.~Lei, S.~Liao, and S.~Z. Li.
\newblock Learning face representation from scratch.
\newblock {\em CoRR}, abs/1411.7923, 2014.

\bibitem{DBLP:journals/spl/ZhangZLQ16}
K.~Zhang, Z.~Zhang, Z.~Li, and Y.~Qiao.
\newblock Joint face detection and alignment using multitask cascaded
  convolutional networks.
\newblock {\em {IEEE} Signal Process. Lett.}, 23(10):1499--1503, 2016.

\end{thebibliography}
}

\end{document}